\documentclass{amia}
\usepackage{graphicx}
\usepackage[labelfont=bf]{caption}
\usepackage[superscript,nomove]{cite}
\usepackage{color}
\usepackage{amsmath}
\usepackage{cancel}
\usepackage{pgfplots}
\usepackage{wrapfig}
\usepackage{hyperref}
\pgfplotsset{width=10cm,compat=1.9}

\begin{document}
\setlength{\abovedisplayskip}{5pt}
\setlength{\belowdisplayskip}{5pt}

\title{Personal Health Knowledge Graphs for Patients}

\author{Nidhi Rastogi, Ph.D.$^{1}$, Mohammed J. Zaki, Ph.D.$^{1}$}

\institutes{
    $^1$Rensselaer Polytechnic Institute, Troy, New York, U.S.A.\\
}

\maketitle

%\noindent{\bf Abstract}

%\textit{Abstract text goes here, justified and in italics.  The abstract would normally be one paragraph long.  See Table 1. for appropriate abstract length by submission type.}

\section*{Abstract}
Existing patient data analytics platforms fail to incorporate information that has context, is personal, and topical to patients. For a recommendation system to give a suitable response to a query or to derive meaningful insights from patient data, it should consider personal information about the patient's health history, including but not limited to their preferences, locations, and life choices that are currently applicable to them. In this review paper, we critique existing literature in this space and also discuss the various research challenges that come with designing, building, and operationalizing a personal health knowledge graph (PHKG) for patients.
\newline
\textbf{Keywords} : \textit{Personal Health Knowledge Graphs, Knowledge Graphs, Diabetes.}

\section{Introduction}
Knowledge Graphs (KG) encode structured information of entities and their relations by capturing information retrieved from several resources. They are represented by a pre-defined ontology that uses different classes and the relationships identified between these classes. KGs confer the ability to search information efficiently, and can help find and utilize patterns in the data for improving clinical outcomes. A few examples of KGs include Google’s Knowledge Graph \cite{Steiner}, sources such as DBpedia\cite{Lehmann}. Publicly accessible KGs have been successfully operationalized to gather insights from medical data sets. However, they do not offer reasoning over observations of daily living (ODLs), which can inform preventative care of chronic health conditions such as diabetes, alzheimer, and asthma. KGs also do not cater to results specific to a given patient. Nonetheless, progress can be made by incorporating patient lifestyle information that is contextual, personal and topical. In this regard, personal health KGs can offer solutions that encourage bringing together medical, social, behavioral, and lifestyle information and grasp nuances of a patient's health to a greater extent. Entities from a personal KG represent the daily tasks and interactions of a specific patient. For example, if a patient were to query a recommendation system for a food recipe, a KG would get called with responses that contain any food recipe recipes. On the other hand, for a query such as $"$What is in the lunch menu at my favorite Indian restaurant?$"$, a recommendation system enabled by a PHKG would return more personalized responses.

\section{Defining Personal Health Knowledge Graphs (PHKG)}
A PHKG represents aggregated multi-modal data that includes all the relevant health-related personal data of a patient by representing it in the form of a structured graph. The data usually comes from various heterogeneous sources like smart phones, survey forms, clinical notes, and other sources that holistically capture a patient's data. There are also different terms used to describe personal KGs. Personalized KGs are curated KGs and are limited by the entities described in the general KGs. Whereas, Personal KGs complement general KGs with additional, personal information about the patient. The entities and their attributes can change with time and so will the associated data. Some of the concepts that can be used to describe a Personal Health Knowledge Graph (PHKG) are described below:
\begin{enumerate}
    \item\textbf{Contextual} - Most of the information captured by a generic KG is not relevant to a patient. A personal health KG, instead, can represent more fluid, contextual, and more rapidly changing information about the patient. Consider a  diabetes patient querying an online health platform that recommends food options to encourage healthy life style. A PHKG is cognizant to the patient's health conditions and nutritional requirements.
    
    \item\textbf{Personal} - A PHKG infers entities and relationships that represent patient interests and information. It evolves with time based on a patient's personal preferences, and interests depending on or relating to the circumstances that form the setting for an event, statement, or idea. A PHKG considers current health condition, health goals, eating habits, food consumption, and also the cultural background of the patient.
    
    \item\textbf{Integrated with existing Knowledge Bases} - A general purpose KG contains prominent information in the form of classes that are \textit{globally} instantiated and accessed by most patients. Hence, the information is described as a class or subclass and instantiated for patients. However, classes or entities that have relevance to very few patients, and are not available in the public domain can be captured by PHKGs. This kind of information is represented by small-sized, structured graphs which are integrated with the larger, general purpose KG while also ensuring that there is entity linking taking place.
\end{enumerate}

\textit{Representing a PHKG} - There is no standard model for
representing a PHKG. So far, published research shows different models,
each based on the use-cases that motivated their formation. For example,
researchers\cite{Balog} recommend a personal KG with a distinctive
spider-web like layout where the patient is the root-node entity. We
infer from the description that the patient's personal information could
be represented by new classes or sub-classes to represent wide variety
of entities, attributes of entities, as well as relationships between
them. Another recommended\cite{Safavi} approach involves decomposing a
large graph into sub-graphs, such that the the nodes within a sub-graph
are highly inter-connected. The identification of these sub-graphs is
significant as they can help uncover unknown modules in such graphs. 

\section{Literature Review}
In this section, we review and provide a critique of various KGs approaches created to extract personal context (or similar) from patient data, especially those suffering from various chronic diseases. While use cases may vary, the intent is to offer a personalized approach for health related recommendations by utilizing small-sized PHKG. Safavi et al.\ \cite{Safavi} describe PHKGs as summaries or $"$mini-KG$'$s$"$ that contain relevant facts about the patient. Patients have limited information capacity, and not all entities are applicable to all patients. However, a limitation of this approach is that the personalized summary is static and contains limited number of entities that are pre-assigned and updated by the system. Updating these entities and relationships is unmanageable with time unless they follow a common graph pattern. Also, the initial step to  create personalized summaries is the patient showing an intent, which comes when they query the recommendation system for the first time. Given the limited memory of smart phones and the processing capabilities, it is can be challenge to create PHKG $'$on-the-fly$'$ with data extracted from heterogeneous sources.

Balog et al.\ \cite{Balog} segregate the purpose of using a KG and a PKG. In a KG (also called public KG), entities have global importance and include resources that are both general purpose and domain specific. They usually miss out on the long tail of entities not prominent enough to have their own Wikipedia articles \cite{Lin} or are non-publicly available facts. However, they define personal knowledge graphs (PKG) as a the portion of the KG that the patient wants to share with the system. While this has privacy protection undercurrents, not all patients are aware enough to know what should and shouldn$'$t be shared. Therefore, it will be ideal to build a hybrid approach where both patient and healthcare professional can collaborate and identify the requirements for the PHKG.

Gyrard et al.\ \cite{Gyrard} consider a patient's personalized knowledge by gathering information from heterogeneous sources such as environmental sensors and web-based data, aggregating, managing, integrating with existing knowledge bases and meaningfully analyzing the collected information. Their kHealth \cite{Sheth} project assists clinicians by aggregating data that belongs to patients of asthma, obesity, and Parkinson's disease. Since the objective of this work is to make recommendations to the clinicians through knowledge gathering and data analysis, it does not support patient-side recommendations input feedback into the daily logs.

Faber et al.\ \cite{Faber} suggest that device memory and computing resources are a limitation and should be considered as essential factors that determines the size and span of the PHKG. Smartphones today have powerful multi-core processors and large memory capacity, so their claim needs further substantiation, and and analysis to understand the bounds of memory and computation requirements.

In order to construct personal knowledge bases, Yen et al. \cite{yen} use text-based life logs shared on social media platforms such as Facebook and Twitter and are captured in the form of tweets or short statements. The purpose of this research is to provide complementary information for recall and retrieval which leads to constructing a PKG. The approach, however, falls short of entity linking with the general knowledge bases. Also, sharing one's daily life events on social media is not necessarily a reliable and consistent source of data. Health information especially may not even be shared by patients with different health conditions. Other daily event journalling too might be sparse, if at all by patients.

\section{Generating a PHKG}
For the most part, a brute force approach has been recommended in published research. Ontologies are created or already exist for global KGs, but not for the much smaller-sized PHKG. In order to generate a PHKG, a quick approach includes inferring patient preferences over a given general purpose KG, and then constructing the patient's PHKG from these preferences. Usually KGs already exist as part of an actual infrastructure that has collected patient data over a period of time. Faber et al.\ \cite{Faber} recommend that once the patient is in, different approaches can be used to dynamically create and instantiate the PHKG. For instance, queries from a patient can be used as input from which other entities and relations of potential interest to the patient can be inferred and call this creating a PHKG \textit{'on-the-fly'}. An interesting, and sometimes useful approach, it does not consider several scenarios. Such as creating relationships and identifying accurate classes for the attributes, and removal of irrelevant entities. Also, they fail to address how individual PHKG will be created or scaled, and what kind of graph structure will be used to model each PHKG.

\section{Challenges}
PHKG is the new research frontier towards making recommendations to patients of various chronic diseases by also incorporating daily life data. Challenges include defining the scope of health and habit related concerns to ensure personal recommendations. Also, validation mechanisms should be devised to ensure that a PHKG effectively captures a patient's personal information over an existing KG. The presence of numerous modalities to (e.g. IoT sensors, smartwatches, apps) will potentially create ingestion, processing, and scalability issues. These questions will help us drive future research work in this area. We conclude that in order to make a research leap in the area of personal health recommendation, several relevant research related questions should be answered.
\newline
\newline 
\textit{ \textbf{Acknowledgement} : This work is supported by IBM Research AI through the AI Horizons Network.}

\bibliographystyle{unsrt}

\end{document}